%% file: ijcai23.tex
\documentclass{article}
\usepackage{ijcai23}

\usepackage{times}
\usepackage{soul}
\usepackage{url}
\usepackage[hidelinks]{hyperref}
\usepackage[utf8]{inputenc}
\usepackage[small]{caption}
\usepackage{graphicx}
\usepackage{amsmath}
\usepackage{amsthm}
\usepackage{booktabs}
\usepackage{algorithm}
\usepackage{algorithmic}
\usepackage[switch]{lineno}
\usepackage{multirow}
\usepackage{subcaption}

\title{LISSNAS: Locality-based Iterative Search Space Shrinkage for Neural Architecture Search}

\author{
Bhavna Gopal\footnotemark[1]$^{1}$
\and
Arjun Sridhar\footnotemark[1]$^{1}$\and
Tunhou Zhang$^1$\And
Yiran Chen$^1$
\affiliations
$^1$Duke University\\
\emails
\{bhavna.gopal, arjun.sridhar, tunhou.zhang, yiran.chen\}@duke.edu
}
\begin{document}

\maketitle
\renewcommand*{\thefootnote}{\fnsymbol{footnote}}
\footnotetext[1]{Co-First Author}
\begin{abstract}
\input{abstract}
\end{abstract}

\section{Introduction}
\input{intro}
\section{Related Work}
\input{relatedwork}
\section{Definitions and Theoretical Motivation}
\input{dtf}
\section{Methods}
\input{methods}
\section{Results}
\input{results}
\section{Conclusion}
\input{conclusion}


\section*{Acknowledgments}
\input{ack}
\section*{Contribution Statement}
Both first authors, Arjun Sridhar and Bhavna Gopal, contributed equally to the production of this work. 
\bibliographystyle{named}
\bibliography{ijcai23}
\appendix
\input{appendix}
\end{document}

%% file: abstract.tex
Search spaces hallmark the advancement of Neural Architecture Search (NAS). Large and complex search spaces with versatile building operators and structures provide more opportunities to brew promising architectures, yet pose severe challenges on efficient exploration and exploitation.
Subsequently, several search space shrinkage methods optimize by selecting a single sub-region that contains some well-performing networks. 
Small performance and efficiency gains are observed with these methods but such techniques leave room for significantly improved search performance and are ineffective at retaining architectural diversity. 
We propose LISSNAS, an automated algorithm that shrinks a large space into a diverse, small search space with SOTA search performance. 
Our approach leverages locality, the relationship between structural and performance similarity, to efficiently extract many pockets of well-performing networks. We showcase our method on an array of search spaces spanning various sizes and datasets. We accentuate the effectiveness of our shrunk spaces when used in one-shot search by achieving the best Top-1 accuracy in two different search spaces. Our method achieves a SOTA Top-1 accuracy of 77.6\% in ImageNet under mobile constraints, best-in-class Kendal-Tau, architectural diversity, and search space size.

%% file: intro.tex
Neural Architecture Search (NAS) \cite{cai2019ofa,liu2019darts,tan2018mnas} has attracted attention by achieving state-of-the-art performance on several tasks. Search space selection serves as a crucial predecessor to NAS by defining the space of architectures being considered during search. Large and complex search spaces have an increased likelihood of containing better-performing candidates. Additionally, technical advancements in operation choices, architecture complexity, and network depth/width have resulted in an increased need for such spaces. However, these large and complex search spaces come with a slew of optimization and efficiency challenges \cite{angle2020}.

\begin{figure}[t]
\begin{center}
    \includegraphics[width=\linewidth]{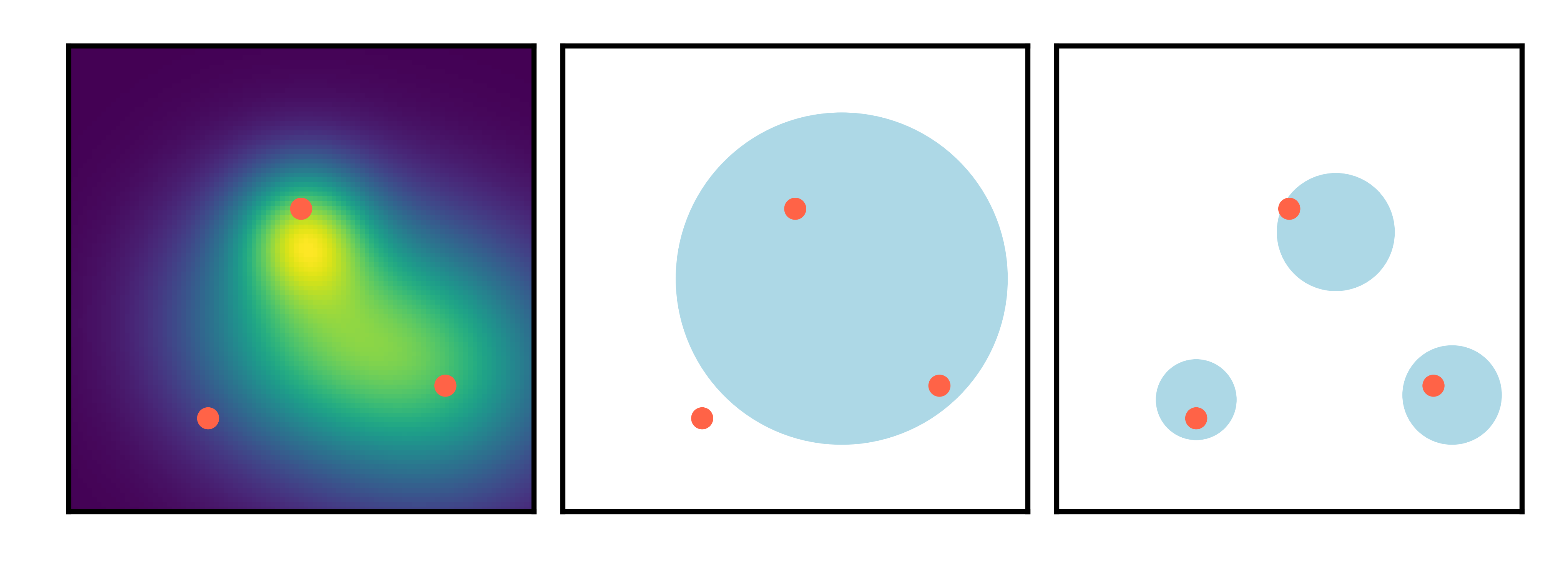}
\end{center}
\caption{The graphs represent a 2-D embedding of a high-dimensional search space. The red dots represent the 3 best-performing networks in the space. {\it Left:} A heatmap with lighter colors representing regions with higher accuracy. {\it Middle:} The blue region represents the shrunk search space produced by other methods. {\it Right:} We produce a smaller search space (indicated by smaller blue area) that is a combination of the 3 regions.}
\label{fig:ss}
\end{figure}

Past efforts have focused on improving NAS search techniques to accommodate increasing search space size and complexity. Nevertheless, search space improvements itself have been relatively underutilized for driving improvement. More recently, research has shown that search space shrinkage algorithms can significantly improve the performance of a variety of NAS methods \cite{angle2020}. As a result, shrinkage methods were proposed to shrink the search space, while retaining some top-performing networks. Most of these approaches shrink the original space to a single region consisting of a relatively higher proportion of well-performing networks. To do so, these algorithms utilize \emph{operational pruning} - either explicitly or implicitly pruning lower-ranked operational choices from the space. Here, ranking and pruning are holistic - all architectures within the space containing the operation contribute to its perceived value. From this standpoint, if an operation is frequently found in poorly-performing architectures then the rank of the operation is likely lower. This results in all networks containing lower-ranked operations being removed from the space. However, as illustrated in Figure \ref{fig:ss}, operational pruning is too aggressive. Inevitably, this technique removes well-performing networks from other regions of the space that typically contain a lower proportion of optimal networks. Hence, we aim to better differentiate and retain optimal architectures throughout the space to improve search performance and alleviate the challenges brought by large search spaces. 

We propose a new paradigm, LISSNAS: Locality-based Iterative Search space Shrinkage for Neural Architecture Search. In contrast to existing approaches, our method comprehensively extracts a smaller search space with an extremely high proportion of diverse, well-performing networks from the original space. This directly results in improved NAS performance. LISSNAS aims to incorporate more well-performing networks, prune poor-performing networks, and maintain diversity across the FLOP range in our search space. 

To accomplish these goals, our algorithm hinges on \emph{architectural pruning} by leveraging locality, the property that structural similarity between architectures correlates to performance similarity. We begin by sampling networks from the original space. Then, using a predictor, we evaluate the samples and leverage locality to construct more clusters of networks that are likely to be well-performing. Eventually, we extract and selectively query pockets of likely well-performing networks in contrast to querying all architectures in the space. We note that even with a predictor, querying all architectures in the space is time-consuming and computationally expensive as spaces can contain upwards of $10^{18}$ architectures. After several automated iterations of refinement, our output, final, search space consists of a combination of well-performing pockets of networks.

 We showcase our method on a vast array of search spaces with various sizes across datasets. We accentuate the effectiveness of our shrunk spaces when used in one-shot search by achieving the best Top-1 accuracy in two different search spaces. Our method achieves a SOTA Top-1 accuracy of 77.6\% in ImageNet under mobile constraints, best-in-class Kendal-Tau, architectural diversity, and search space size. Overall, our contributions are as follows:
\begin{enumerate}
    \item We successively establish locality preservation quantitatively within 3 popular search spaces (ShuffleNetV2, NASBench101, TransNASBench) and 6 datasets (ImageNet, CIFAR10, 4 datasets in TransNASBench - scene, object detection, jigsaw, and semantic segmentation). 
    \item We introduce a new metric for quantitative comparison across different shrinkage algorithms, the \emph{shrink index}. This index captures the improved probability of sampling a well performing network in our shrunk search space. 
    \item We demonstrate that LISSNAS improves overall NAS performance (Top-1) in addition to several search space quality metrics. We efficiently achieve SOTA Top-1 accuracy of 77.6\% on ImageNet under mobile constraints.  
\end{enumerate}

%% file: relatedwork.tex
\subsection{General NAS}
Designing a CNN poses a challenge to researchers looking to optimize for many constraints such as latency, model size, and accuracy. Recent work in NAS attempts to automate the process of designing and discovering CNN architectures that meet these constraints. In 2017, NAS with Reinforcement Learning \cite{zoph2017neural} introduced the concept of NAS utilizing an RL-based controller that iteratively proposed and trained candidates. Training the controller and candidates was computationally exorbitant. Since then, several utilized genetic algorithms \cite{real2019regularized,real2017large,xue2021multi}, Bayesian optimization \cite{pmlr-v28-bergstra13,mendoza2016towards}, and predictors \cite{wen2020neural,white2021powerful,wei2022npenas}. 

The next significant milestone was the introduction of weight sharing through one-shot networks, also referred to as supernetworks. SMASH \cite{brock2017smash} employs a One-Shot Network that uses a set of shared weights to train many candidates simultaneously. At the time, this approach significantly reduced training time but limited architecture exploration, impacting performance. To alleviate this limitation, OFA \cite{cai2019ofa}, Attentive NAS \cite{wang2020attentivenas}, and several other works train these networks through novel path sampling methods and a host of optimization techniques \cite{pmlr-v139-wang21i,luo2020balanced,luo2019understanding,pham2018efficient,xie2018snas,zhang2020you}. Overall, one-shot networks proved to be efficient and resulted in SOTA NAS performances across a variety of tasks.
 
\subsection{Search Space}
\noindent \textbf{Search Space Shrinkage Motivation.} 
Angle-Based Search (ABS)~\cite{angle2020} investigated behaviors of NAS methods on various shrunk search spaces. The results show that shrunk search spaces have the potential to enhance existing NAS algorithms across many different search methods. Neural Search Space Evolution (NSE) \cite{evolv_ss} presents the drawbacks of large search spaces and acknowledges the importance of the size and design of the search space in supernet training. Additionally, several others \cite{Yu_ss,Zhang_ss} have shown that enlarging a search space typically results in significant accuracy loss for the final searched architecture. Lastly,  SPOS \cite{guo2019singlepath} determined that search spaces of reduced cardinality could achieve higher correlations.

\noindent \textbf{Search Space Shrinkage.}
Several methods have attempted to shrink the search space while preserving the top-performing networks \cite{angle2020,padnas2020,You2020GreedyNASTF,greedyNAS2022,radosavovic2020designing,tyagi2020visual}. GreedyNAS \cite{You2020GreedyNASTF} proposes an automated NAS optimization strategy that greedily samples well-performing networks during training. NSE \cite{evolv_ss} simply segments the search space and trains in a guided way similar to GreedyNAS. These works fundamentally use a controller to efficiently sample networks in training and evaluation without truly constructing a new search space. ABS drops the lowest-performing operation choices until the search space is shrunk to the desired size. Similar to ABS, PADNAS \cite{padnas2020} also prunes operations but does so on a layer-wise basis.

In 2020, several manual (human-dependent) techniques were proposed to build optimal search spaces. Designing Network Design Spaces (DDS) \cite{designing2019,radosavovic2020designing} utilizes statistical analysis to assist researchers with iterative search space shrinkage. Additionally, Visual Steering for One-Shot \cite{tyagi2020visual}, uses promising structures proposed by specialists to build a search space and eventually converges at the optimal network. Both of these approaches are time-consuming and require iterations of domain expert input for success. 

%% file: dtf.tex
\subsection{Definitions, Instantiations, and Assumptions}
Our work utilizes specialized terminology such as locality, search space, model distribution, edit distance, isomorphism, and neighbors. For clarity on the definition of these terms please refer to Appendix B.

This work makes the following assumptions:

\begin{enumerate}
    \item All potential changes that can be made to a network in a search space are of equal value (Appendix G). 
    \item To determine the statistics of our final shrunk search spaces, we assume that a distribution of networks is more indicative than point sampling networks \cite{designing2019}. 
\end{enumerate}

\subsection{Benchmark}
We investigate locality and our algorithm’s performance on NASBench101, TransNASBench Macro, NASBench301, and ShuffleNetV2. For a more detailed description of these search spaces and respective datasets, please reference the Appendix C.  

\subsection{Theoretical Motivation}
Generally speaking, NAS is more optimal if a comparable (or better) top-1 accuracy is obtained from search in less time. We now explain how a smaller search space with a higher density of well-performing networks increases NAS search performance and efficiency (optimality). To do so, we use the example of training a supernet. 

Most large search spaces contain some proportion of weak-performing architectures. In a standard supernet training method, training involves sampling architectures from the search space. Let us assume that we define our search space as $A$, and that $A$ can be cleanly partitioned into $A_{good}$ and $A_{bad}$ \cite{greedyNAS2022}. These partitions refer to potentially optimal and suboptimal networks respectively. Definitionally: 
\begin{equation}
    \resizebox{.91\linewidth}{!}{$
            \displaystyle
            A = A_{good} \cup A_{bad}, A_{good} \cap A_{bad} = \emptyset
        $}
\end{equation}
\newtheorem{theorem}{Theorem}

Assume that we are given an algorithm $F$ to remove relatively poor-performing architectures from a search space. Since $F(A)=A_{shrunk}$, the shrunk search space has reduced cardinality as well as contains a higher proportion of optimal networks. During training, the supernet will now sample more optimal architectures at an increased frequency. Firstly, since the cardinality of the search space decreases, the training efficiency improves. Secondly, an increased proportion of optimal networks in $A_{shrunk}$ enables the supernet to focus training and searching on superior regions of the search space. This results in better performance ranking. 

It follows that we aim to shrink and improve our search space by pruning networks from $A_{bad}$. Unfortunately, we are typically unaware of the exact partition in A and are hence unable to precisely determine which networks to prune, without ground truth accuracies. 

%

%% file: methods.tex
\subsection{Assumptions}
To reiterate, the goal of the proposed pipeline is to shrink a large search space by identifying and extracting well-performing search space subsets. We begin by stating the assumptions of our method and empirically verify locality preservation. We then utilize our observations to develop our method and obtain results. Lastly, to bolster our approach, we also perform some ablation experiments.
\begin{figure}[h]
\begin{center}
    \begin{subfigure}{.167\textwidth}
      \centering
      \includegraphics[width=\linewidth]{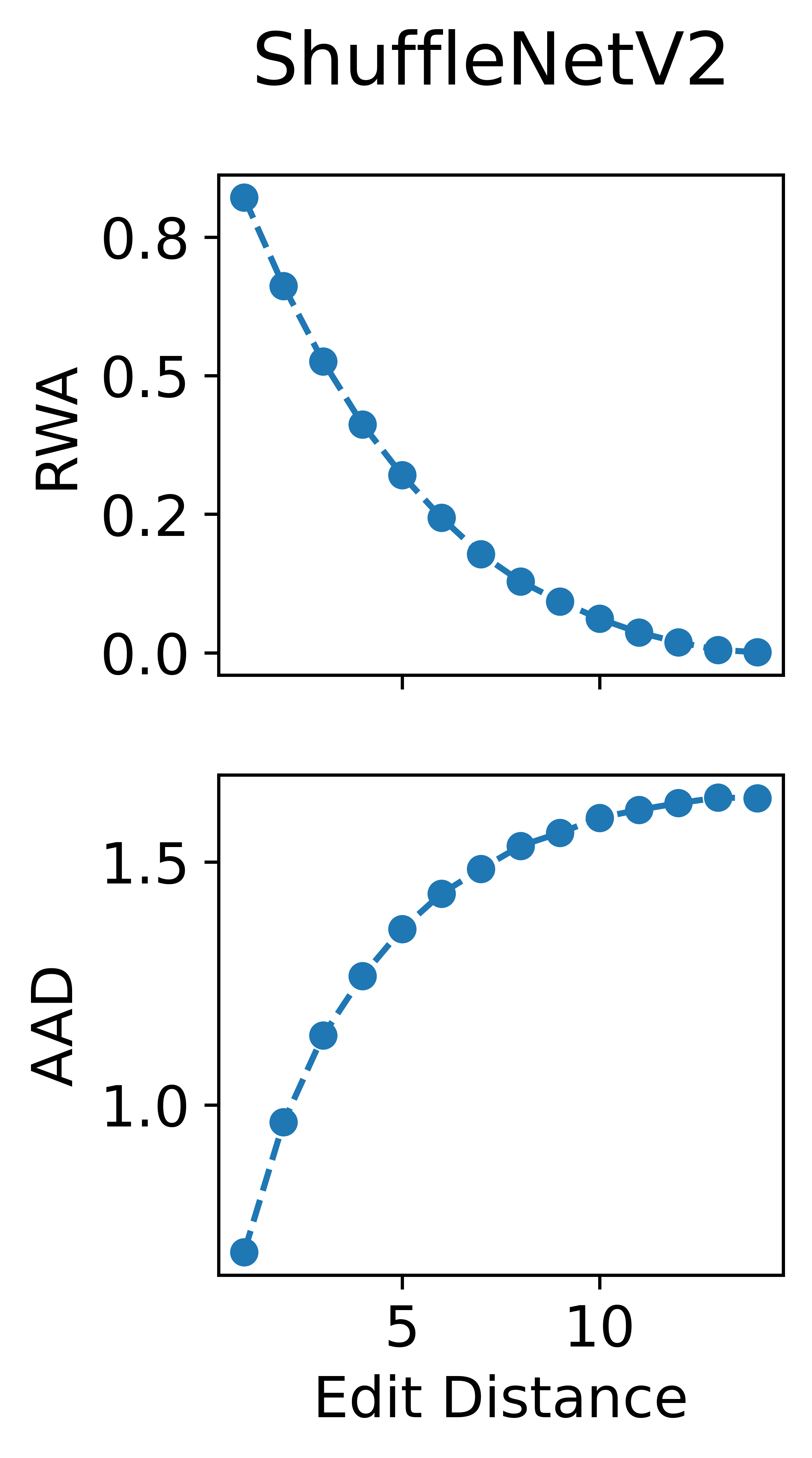}
    \end{subfigure}
    \begin{subfigure}{.15\textwidth}
      \centering
      \includegraphics[width=\linewidth]{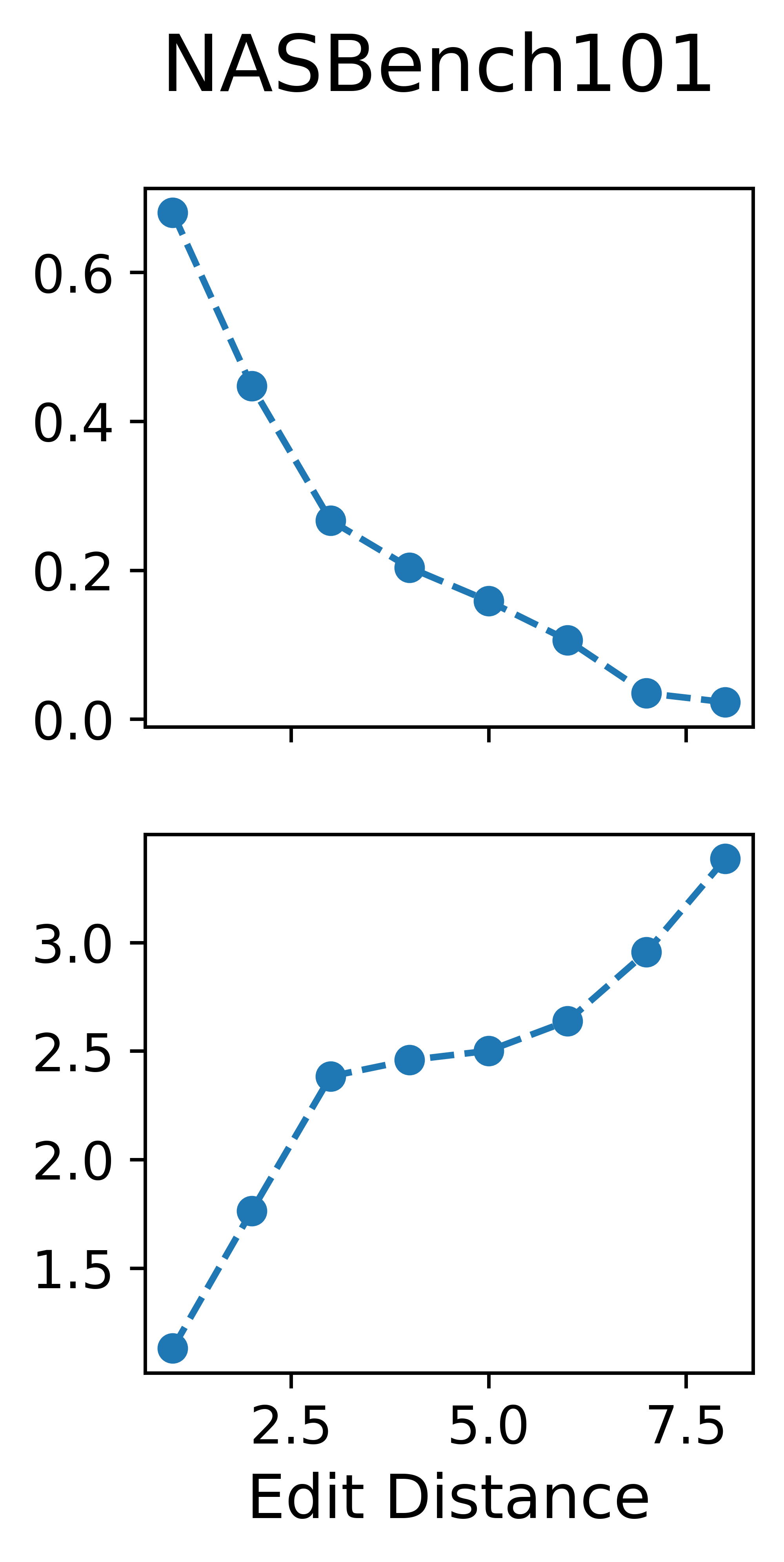}
    \end{subfigure}
    \begin{subfigure}{.15\textwidth}
      \centering
      \includegraphics[width=\linewidth]{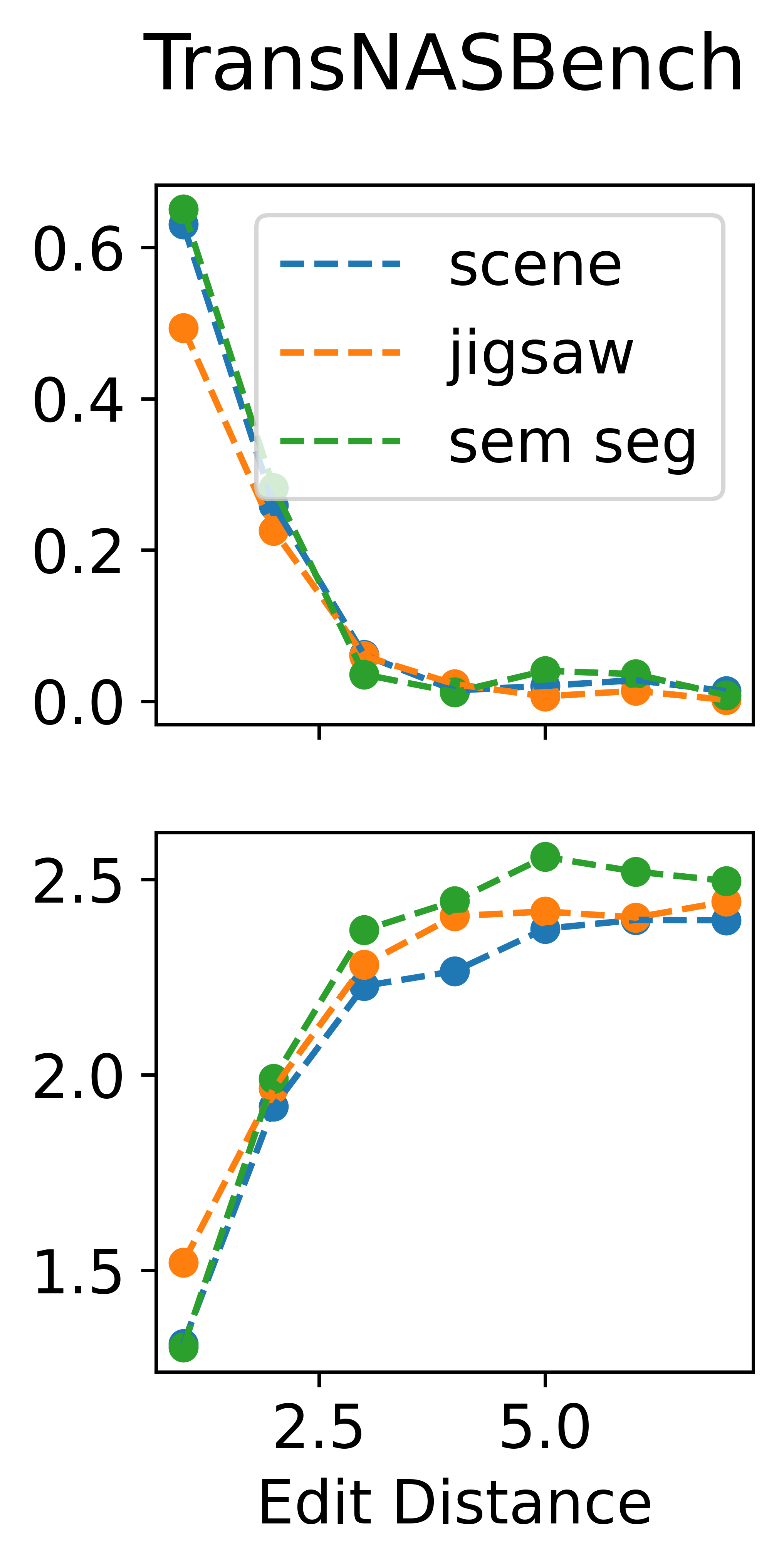}
    \end{subfigure}
\end{center}
\caption{The top row of graphs demonstrate RWA. The correlation drop-off is steep and the correlations are weak ($<.3$) at one-third of the total edit distance available. The bottom row of graphs show the AAD. The trends are consistent across all search spaces and tasks.}
\label{fig:rwa}
\end{figure}
\subsection{Locality Preservation}
Before locality exploitation, we empirically verify that locality is preserved across our search space types, sizes, and tasks. To do so, we illustrate Random Walk Autocorrelation (RWA) and compute the averaged Absolute Accuracy Difference (AAD) between generated neighbors and original networks. 

To compute the AAD and RWA, we define our metric of similarity between networks as the \emph{edit distance} between networks - the minimum number of changes required to change one network into another.  
Traditionally, computing the edit distance between two networks is computationally expensive. To alleviate the burden of calculating pairwise edit distances, we formulate our problem in terms of neighbor generation as opposed to edit distance computation, i.e: instead of computing the edit distance between two sampled networks, our method randomly samples a network and makes changes (up to a certain edit distance) to obtain another network. The generated network is referred to as a \emph{neighbor}.

Firstly, we demonstrate the RWA between the accuracies of the sampled network and the generated neighbor \cite{ying2019nasbench}. RWA is defined as the autocorrelation of the accuracies of networks as a random walk of changes is made. As the number of changes increases the correlation of accuracies with the original network should decrease. Our observations match our hypothesis - the RWA decreases as the number of changes made to a network increases as shown in Figure \ref{fig:rwa}.  

Our RWA results are corroborated by determining the AAD between networks at increasing edit distances. We observe that as the number of changes made to a network increases, the accuracy gap (between the original and the generated neighbor) increases. This directly demonstrates locality by accentuating that networks that have more structural similarity perform similarly. Notably, all of our search spaces preserve locality and we summarize our observations in Figure \ref{fig:rwa}. Our observations give rise to the premise of our algorithm. If we find a relatively well-performing network within a search space, we generate neighbor networks to add to our new shrunk space. The next section outlines the proposed locality-guided shrinkage algorithm. 

\begin{figure}[t]
    \centering
    \includegraphics[width=\linewidth]{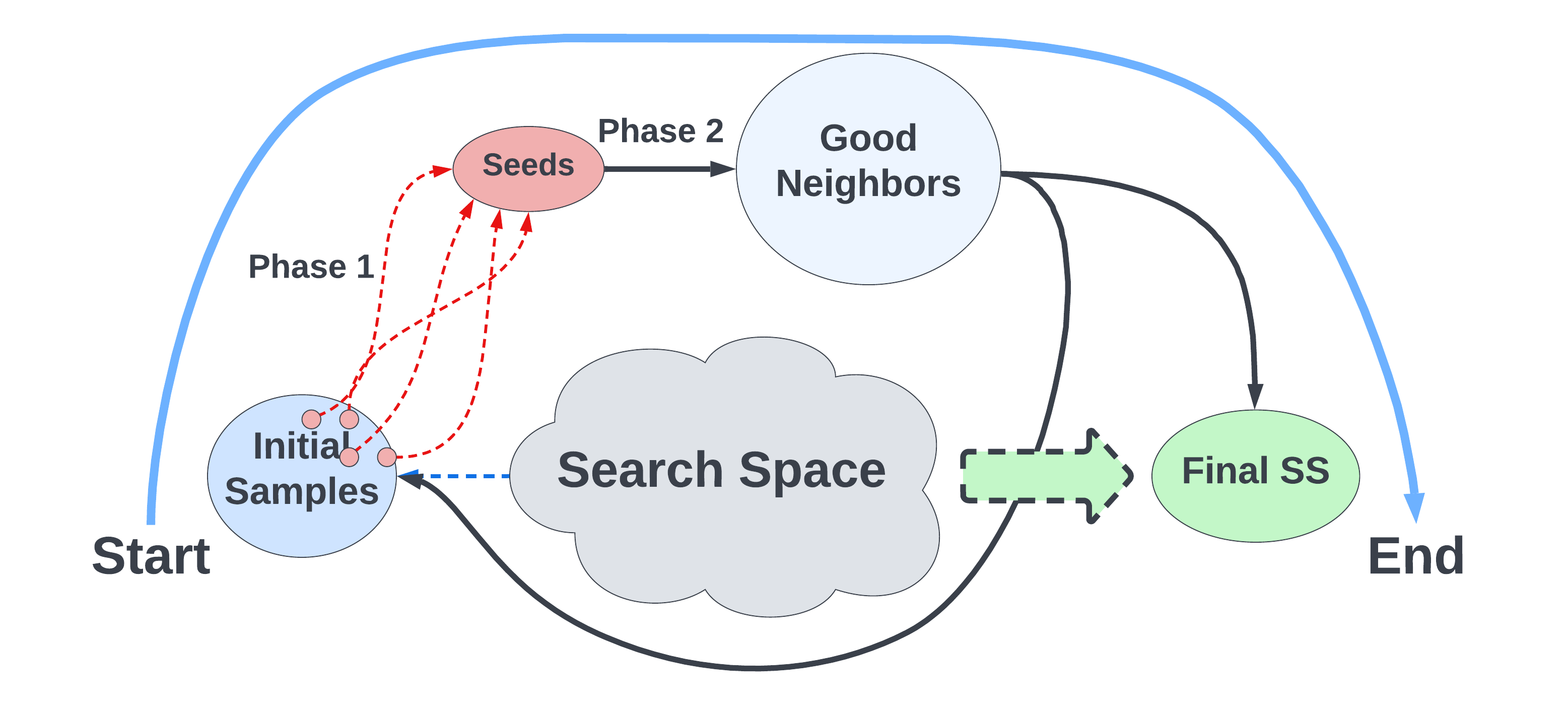}
    \caption{LISSNAS begins by sampling networks from the space. Next in Phase 1, the best samples are taken as the seed networks. These seeds are used in Phase 2 to generate good neighbors through locality exploitation. Finally, we determine whether to continue sampling and refining the space or terminate with our current shrunk search space.}
    \label{fig:lissnas}
\end{figure}

\subsection{Locality-based Iterative Search Space Shrinkage}
Figure~\ref{fig:lissnas} appropriately highlights the pipeline of our shrinkage approach. We simplify our approach to 2 main stages: seed search and neighbor generation. 
\begin{enumerate}
    \item {\bf Seed Search:} search for well-performing networks within a given search space
    \begin{enumerate}
        \item Our initial networks are obtained by randomly sampling the original space. Note that we sample at least 1000 networks from each of our search spaces based on our understanding of network distributions.
        \item Since we are unaware of the ground truth accuracies of our initial samples, an accuracy predictor determines performance. Similar to NASBench301, we utilize XGBBoost as our accuracy predictor. NASBench301 outlines the effectiveness of this predictor in modeling performances while providing low-cost inferences.
        \item Based on our predicted accuracies, we select the top-performing networks and refer to these as \emph{seeds}.
    \end{enumerate}
    \item {\bf Neighbor Generation:} generate neighbors for our seeds
    \begin{enumerate}
        \item For each seed, we generate several neighbor networks at small edit distances. To do so, we determine a threshold edit distance. This is the edit distance up to which we can generate neighbors for each seed. Through our analysis of RWA, we found a good threshold is 1/3 of the total available edit distance as shown in Figure~\ref{fig:rwa}. Note: Evaluating all architectures is prohibitively expensive. We exploit locality to generate neighbors in order to guide search. Details on this are in the ablation study section (Appendix H).   
        \item The seed networks along with their neighbors constitute the shrunk search space with a higher proportion of optimal nets. We compute the mean accuracy of the constructed search space and perform more iterations if the average accuracy increases. This entails going back to the seed selection step and determining new seeds. Since locality is preserved within the space, a neighbor is likely to perform similarly to its seed. 
        \item If the average accuracy decreases or plateaus, we output our previous iteration's search space. From this perspective, our algorithm shrinks the search space to the maximum extent possible.
    \end{enumerate}
\end{enumerate}

\subsection{Optimality Verification}
Once we obtain the output shrunk search space, we verify its performance. Primarily, our algorithm hinges on reduced cardinality and an increased proportion of well-performing networks. We verify these metrics in live time (as the algorithm runs) by ensuring that the size of the search space reduces while the average search space accuracy increases. However, for easier comprehension and generalizability, we would like to encapsulate these baseline metrics through quantitative parameters. Given the lack of standardized metrics (or approaches) for search space quality evaluation, how do we verify that a search space has shrunk effectively?

 We first clarify our theoretical definition of {\it effective shrinkage}. Understanding this term is crucial for a preliminary indication of the improvement and effectiveness of our algorithm across iterations. We consider a shrinkage algorithm to be both \emph{optimal} and \emph{effective} if these conditions are met: 1) the cardinality of the search space decreases with each iteration; and 2) the shrunk search space has a higher probability of containing well-performing networks than the original. The second condition implies that the probability of the search space containing a well-performing network increases as the iterations increase until algorithm termination.

Secondly, to verify that our search space has shrunk effectively, we determine the quality of our output shrunk search spaces using a myriad of other metrics and classical statistics. However, in terms of our baseline conditions, our first condition is easily quantified as the difference in size between the output shrunk search space and the original. For the second condition, we define the \emph{shrink index}: 

\noindent \textbf{The Shrink Index.}
If we uniformly sample a network $n$ from $A$, then let's say that the probability that $ n \in  A_{good}$ is $p = |A_{good}|/|A|$. Likewise, the probability that  $ n \in A_{bad}$ is then $1 - p$. If we proceed to sample multiple networks independently and based on the results of a binomial distribution, we have the theorem:

\begin{theorem}
    If n paths are sampled uniformly i.i.d. from A, then it holds that at least k (k $\leq$ n) paths are from $A_{good}$ with probability 
    \[ \sum_{i=k}^{n} \binom{i}{n} p^j (1-p )^{(n-i)},\] where $p=|A_{good}|/|A|.$ 
\end{theorem}

Conservatively, we determined that we would like our search space to have at least 20\% of networks from $A_{good}$. In addition, we can estimate $p$ for a search space by querying a small subset networks. When using the shrink index for comparison, the threshold accuracy  used to generate $p$ should be kept constant.

By using this threshold and $p$, we compute the $P_{good}$ (probability of sampling a good network) initially for the entire initial search space, the final output search space, and at all intermediate iterations. An increase in this probability across iterations implies that the proportion of networks from $A_{good}$ is increasing in the current search space and, as a result, increasing our odds of sampling a well-performing network. We term the difference between the initial probability and the probability in our shrunk search space as the \emph{shrink index} (s.i) as seen in equation 2: 
\begin{equation}
s.i = P(good\mid init\_ss) - P(good\mid shrunk\_ss)
\end{equation}
Intuitively, the shrink index is dependent on the initial search space as well as the dataset of consideration. Shrink indices indicate the increase in the probability of sampling a well-performing network and serve as a good comparison metric across different shrinkage approaches. Generally, higher indices indicate more optimal shrinkage. We note that the shrink index can only be used for comparing algorithms on the same dataset and search space. To illustrate this, consider that we have search space A which is reduced to A*, and search space B which is reduced to B*. Assume the shrink index of search space A* is 0.5 and the shrink index of search space B* is 0.2. However, the actual probability of sampling a good network from A goes from 20\% to 70\% in A* in contrast to B which goes from 70\% to 90\% in B*. In this scenario, while the shrink index of A* is higher, search space B* is likely to be a better search space of choice. This example highlights the importance of using shrink index to compare algorithms on the same starting search space since the initial concentration of networks is important. As shown in Table \ref{table:shrinkind}, our shrink index is significantly higher than other works when compared on ShuffleNetV2. Other search spaces also follow this trend.  


\begin{table}[h]
\begin{center}
\begin{tabular}{ll}
\hline
\noalign{\smallskip}
GreedyNAS     & 0.3    \\
ABS          & 0.1     \\
NSE           & 0.2    \\
PADNAS        & 0.1    \\
\hline
\noalign{\smallskip}
\textbf{Ours} & \textbf{0.4} \\
\hline
\end{tabular}
\caption{Shrink Index in ShuffleNetV2 Search Space on ImageNet dataset}
\label{table:shrinkind}
\end{center}
\end{table}

\subsection{NAS Methodology}
Now that we have a shrunk search space with a verifiable proportion of well-performing networks, we proceed to determine the effectiveness of our shrunk spaces on search performance. Starting with the ImageNet under mobile constraints space, we shrank the FairNAS space using LISSNAS in combination with an XGBoost predictor. From the shrunk search space, we proceeded to run one-shot NAS following the same scheme as K-shot \cite{kshot} with several minor optimizations for hardware. For comparisons against other shrinkage methods, we again utilize a predictor in combination with one-shot NAS to report both Pearson Correlation and Top-1 accuracy. Special care is taken to ensure that all works that are compared receive fair treatment. We use the same predictor for shrinking the search spaces using different methods and then train the one-shot network using the same hyperparameter optimization scheme.

%% file: results.tex

We now present, evaluate, and discuss extensive experimental results to show the effectiveness of LISSNAS. We begin by evaluating NAS performance on ImageNet under mobile constraints before directly comparing with other search space shrinkage works. Finally, we compare the search space statistics, size, and architectural diversity. Like many works in NAS, shrinkage methods are implemented in different search spaces making direct comparison challenging. In the following sections, we implement, optimize, and tune competing methods in order to compare our experimental results.


%

\subsection{ImageNet Classification}
ImageNet under mobile constraints serves as a canonical benchmark for all NAS techniques. As such, we are able to compare not only search space shrinkage techniques but also modern general NAS works. For a fair comparison, we elect to only include standard methods on CNN architectures without pretraining, knowledge distillation, or attention-based mechanisms. We note that Top-1 performance comparisons are feasible despite methods using other search spaces. In Table \ref{table:imagenet}, we see that LISSNAS achieves SOTA accuracy. To obtain these results, we ran LISSNAS on the FairNAS search space containing $6^{19}$ architectures. With this size, querying every architecture becomes unfeasible even with fast inferences. Through leveraging locality, we are able to search this space in only 4 GPU days. 

\begin{table}[t]
\begin{center}
\begin{tabular}{llc}
\hline\noalign{\smallskip}
Model & Flops/Params & Top-1  \\ 
\noalign{\smallskip}
\hline
\noalign{\smallskip}
MobileNetV3* \nocite{mobilenetv3} & 219M/5.4M & 74.0 \\
NASNet-A \nocite{zoph2018learning} & 564M/5.3M & 74.0 \\ 
ShuffleNetV2 \nocite{zhang2017shufflenet} & 299M/3.5M & 72.6 \\ 
SPOS* \nocite{guo2019singlepath} & 319M/3.3M & 74.3 \\ 
OFA \nocite{cai2019ofa} & 230M/- & 76.9 \\
FBNet-C \nocite{fbnet} & 375M/5.5M & 74.9 \\
FairNAS-C \nocite{fairnas} & 325M/5.6M & 76.7 \\
K-shot-NAS-B \nocite{kshot} & 332M/6.2M & 77.2 \\
GreedyNASv2-S \nocite{greedyNAS2022} & 324M/5.7M & 77.5 \\
\hline
\noalign{\smallskip}
DDS \nocite{angle2020} & 400M/4.3M & 74.1 \\
ABS \nocite{angle2020} & 472M/- & 75.9 \\
NSENet \nocite{angle2020} & 330M/4.6M & 75.5 \\
PAD-NAS-S-B \nocite{padnas2020} & 315M/4.1M & 74.9 \\
{\bf LISSNAS} & {\bf 329M/6.1M} & {\bf 77.6} \\ 
\hline
\end{tabular}
\caption{ImageNet under mobile constraints. * Verified reproducible}
\label{table:imagenet}
\end{center}
\end{table}

\noindent \textbf{Comparison Against Shrinkage Works.}
We now directly compare the performance of our method to other shrinkage approaches on the same search spaces - NASBench101 and ShuffleNetV2 search space. To do so, we optimized and implemented these approaches before running one-shot NAS for search. NASBench101 and ShuffleNetV2 search spaces vary greatly in size and complexity  - enabling diverse and fair comparison. 
\begin{table}[t]
\begin{center}

\begin{tabular}{lcccc}
\hline\noalign{\smallskip}
\multicolumn{1}{c}{Search Space} & \multicolumn{2}{c}{NASBench101} & \multicolumn{2}{c}{ShuffleNetV2 } \\
    & Pearson & Top-1  & Pearson & Top-1\\
\noalign{\smallskip}
\hline
\noalign{\smallskip}
Full Search Space  &  0.467  &  91.0 &  0.334  &  70.0  \\
ABS & 0.58 & 93.0 &  0.397  &  72.8 \\
NSE & 0.56 & 92.1 &  0.406  &  73.0\\
PADNAS & 0.59 & 92.4 &  0.355  &  69.9\\
{\bf LISSNAS} &  {\bf 0.659}  &  {\bf 93.4}  &  {\bf 0.429}  &  {\bf 74.6}\\
\hline
\end{tabular}
\caption{Search Space Shrinkage NAS Performance}
\label{table:NASBenchOneShot}
\end{center}
\end{table}

 In Table \ref{table:NASBenchOneShot}, we display the Pearson correlation coefficients and Top-1 performances for each of the methods. Our search space has a significantly higher Pearson correlation in both search spaces demonstrating the value of training a supernet on our shrunk search space compared to other techniques. We repeated our experiments  and computed standard deviations across runs and found all results presented to be statistically significant. 

 Lastly, in both search spaces, LISSNAS also produces the highest Top-1 accuracies. For NASBench, LISSNAS is able to find the best network in the space. We note that this network is not present in the shrunk search space produced by ABS, NSE, or PADNAS. 

\subsection{Search Space Statistics} \label{sss}

\begin{table}[h]
\begin{center}
\begin{tabular}{lccc}
\hline\noalign{\smallskip}
Search Space & Size & Avg Acc & Max Acc \\ 
\noalign{\smallskip}
\hline
\noalign{\smallskip}
Full SS - & \multirow{2}{*}{423K} & \multirow{2}{*}{90.4} & \multirow{2}{*}{94.6} \\
NASBench101 & & & \\
ABS & 100K & 80.2 & 93.2 \\
NSE & 25K & 90.1 & 93.6 \\
PADNAS & 15K & 90.4 & 94.1 \\
{\bf LISSNAS } & 13K & {\bf 92.1} & 94.6 \\
\hline
\noalign{\smallskip}
Full SS - & \multirow{2}{*}{1T} & \multirow{2}{*}{64.7} & \multirow{2}{*}{75.6} \\
ShuffleNetV2 & & & \\
ABS & 100K & 65.1 & 74.9 \\
NSE & 1M & 64.7 & 75.2 \\
PADNAS & 1M & 66.2 & 73.1 \\
{\bf LISSNAS } & 1M & {\bf 67.5} & 75.6 \\
\hline
\end{tabular}
\caption{Shrunk Search Space Statistics for NASBench101 and ShuffleNetV2}
\label{table:ss_metrics}
\end{center}
\end{table}

Aside from search performance metrics, statistical measures provide meaningful insight into the quality of our search space. We compare our output search space to the output search spaces produced by other shrinkage algorithms using these metrics. In Table \ref{table:ss_metrics}, we accentuate that both the average accuracy and maximum accuracy of our output spaces are the highest, indicating an improvement in the number of well-performing networks. This also indicates a higher proportion of well-performing networks, mirroring the results from our shrink index. 

\begin{figure}[t]
    \centering
    \includegraphics[width=\linewidth]{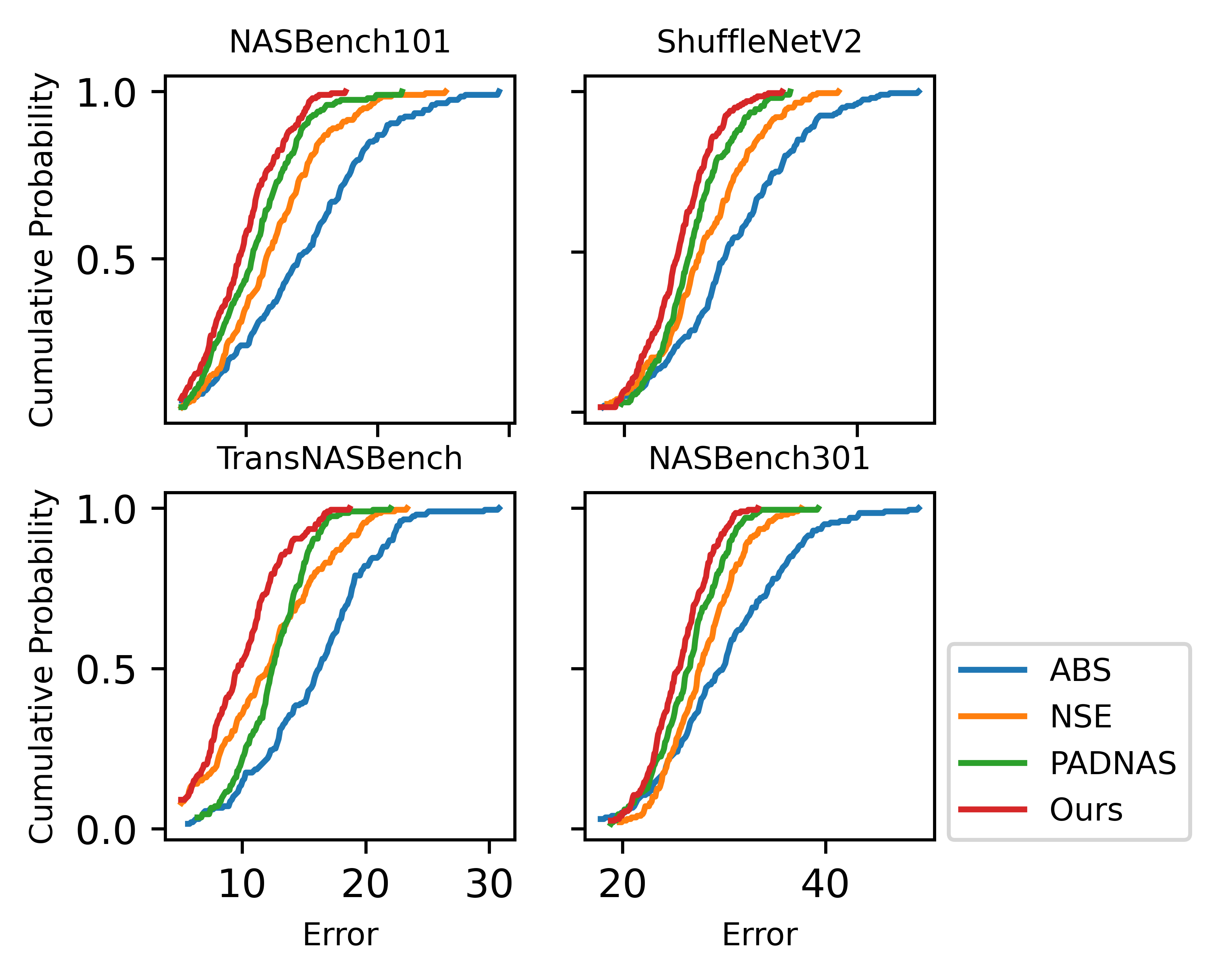}
    \caption{Error EDF captures the quality of a search space. Our algorithm maintains a concentration of better networks across the board.}
    \label{fig:edf}
\end{figure}

Additionally, DDS \cite{designing2019}, demonstrated the effectiveness of error Empirical Distribution Functions (EDF) in evaluating search spaces. Given this prior, we generated error EDFs based on the distribution of networks in our output space, the baseline search space, and the output search spaces of other shrinkage algorithms. These results are presented in Table \ref{fig:edf}. By performing the Kolmogorov-Smirnov (KS) test \cite{radosavovic2020designing}, we established statistically significant improvement with our approach. Notably, this is the case even between our approach and PADNAS which have more visually similar curves. Furthermore, search spaces with a greater area under the curve (AUC) typically have greater concentrations of good networks. Based on this, we can infer that our search space likely contains a larger proportion of good models than the other spaces as our curve covers a larger area. 

\subsection{Cardinality and Extent of Shrinkage}

We begin by comparing our method to ABS and DDS. These methods were implemented on the MobileNet and AnyNetX space respectively. In Table \ref{table:sizecomp}, we compare the cardinality/size and factor of reduction across algorithms. We reiterate that smaller search spaces are feasible solutions for optimization and efficiency issues in NAS. The illustrated reduction factors describe the extent of shrinkage achieved relative to the initial size of the search space. 
Note that DDS is manual design shrinkage method performed by experts and proved difficult to beat. However, our search space offers fairly close cardinality while being generated without expert input. 

Table \ref{table:size} shows the effectiveness of our algorithm on a range of search spaces of varying initial sizes. From this table, we observe that larger search spaces are pruned more as indicated by larger reduction factors. We attribute this to the disparity between the cardinality of well-performing networks in the space and cardinality of the entire space. In more detail, the number of networks that lie in the region next to the Pareto frontier (region of well-performing networks) is much smaller than the overall number of nets in the space. Large and small search spaces have a similar number of networks that lie in the Pareto frontier. However, due to increased cardinality, larger spaces can be pruned to a greater degree, leading to higher reduction factors. 

\begin{table}[t]
\begin{center}
\begin{tabular}{llcc}
\noalign{\smallskip}
\hline
\noalign{\smallskip}
    &          & SS Size & Reduction Factor \\ \hline \noalign{\smallskip}
\multirow{3}{*}{MobileNet}  & Original          & $10^{12}$         & N/A              \\
                            & ABS               & $10^{5}$          & $10^{7} \times$  \\
                            & {\bf LISSNAS}     & $10^{6}$          & $10^{6} \times$  \\ 
\hline \noalign{\smallskip}
\multirow{3}{*}{AnyNetX}    & Original          & $10^{18}$         & N/A              \\
                            & DDS               & $10^{8}$          & $10^{10} \times$   \\
                            & {\bf LISSNAS}     & $10^9$            & $10^{11} \times$  \\
\hline
\end{tabular}
\caption{Size Reduction Comparison}
\label{table:sizecomp}  
\end{center}
\end{table}

\begin{table}[t]  
\begin{tabular}{lccc}
\hline \noalign{\smallskip}
              & \multicolumn{1}{l}{Full SS Size} & \multicolumn{1}{l}{\bf Our Size} & \multicolumn{1}{l}{Reduction} \\ 
\hline \noalign{\smallskip}
TransNASBench   & $7.3\text{K}$     & $1\text{K}$   & $7\times$         \\
\hline \noalign{\smallskip}
NASBench101     & $423\text{K}$     & $13\text{K}$  & $33\times$        \\ 
\hline \noalign{\smallskip}
ShuffleNetV2    & $4^{20}$          & $10^6$        & $10^6\times$      \\ 
\hline \noalign{\smallskip}
NASBench301     & $10^{18}$         & $10^7$        & $10^{11}\times$   \\ 
\hline \noalign{\smallskip}
\end{tabular}
\caption{LISSNAS Size Reduction}
\label{table:size}  
\end{table} 

\subsection{Architectural Diversity} \label{ad}
Lastly, while size provides a good baseline for evaluating our algorithm, exploring diversity is equally important. Multi-objective NAS relies on maintaining the same FLOP/Param range of the original space. As an example, deploying in a mobile setting requires smaller models whereas cloud GPUs with powerful processors utilize large and complex architectures to improve performance. Unexpectedly, we observed that naive search space shrinkage methods tend to drop the lower-performing architectures without considering the need for architectural diversity. From this perspective, we ensured that our algorithm covers the full range of architectures spanned by the original space. 

\begin{figure}[h]
\begin{center}
    \begin{subfigure}{.23\textwidth}
      \centering
      \includegraphics[width=\linewidth]{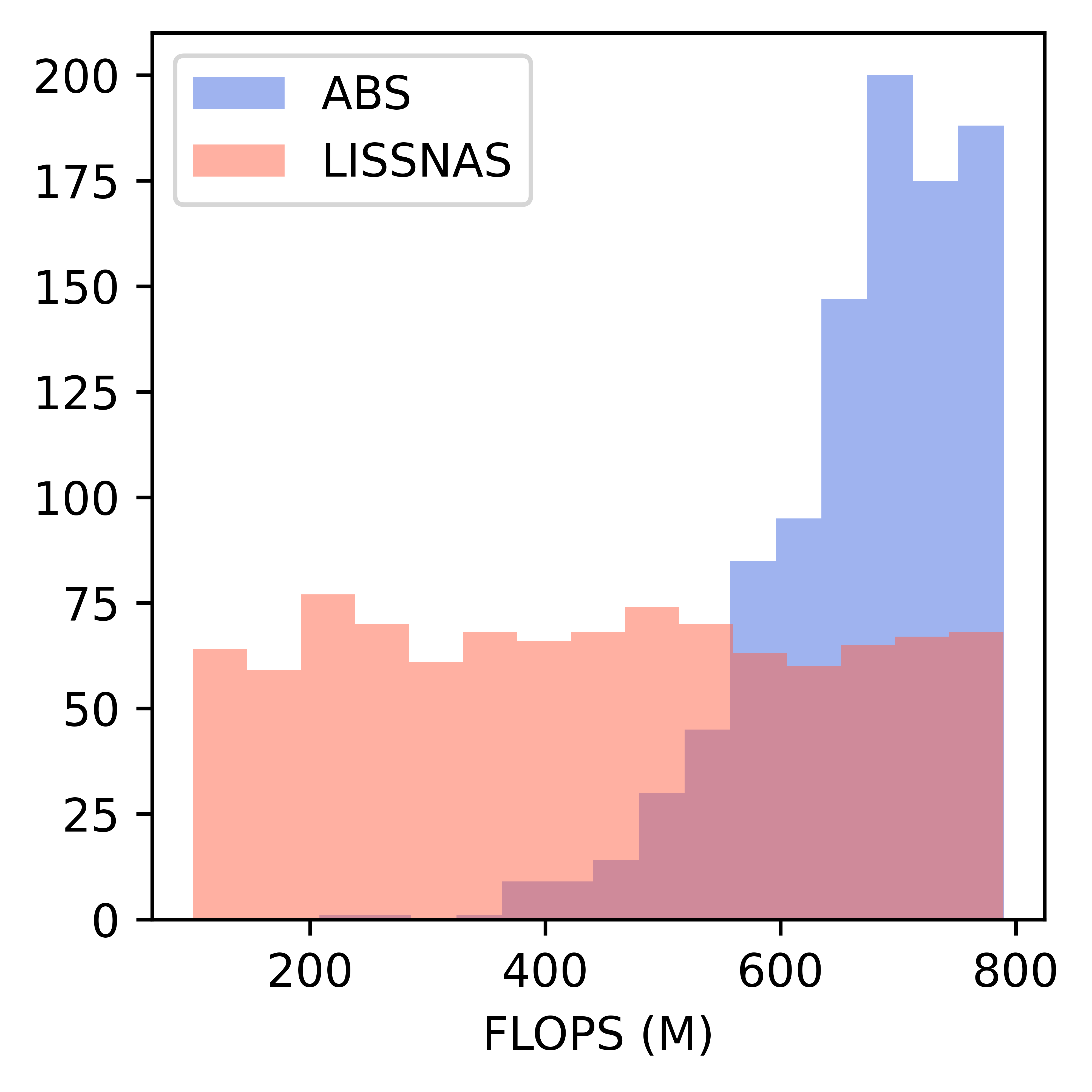}
      \label{fig:flopshist}
    \end{subfigure}%
    \begin{subfigure}{.23\textwidth}
      \centering
      \includegraphics[width=\linewidth]{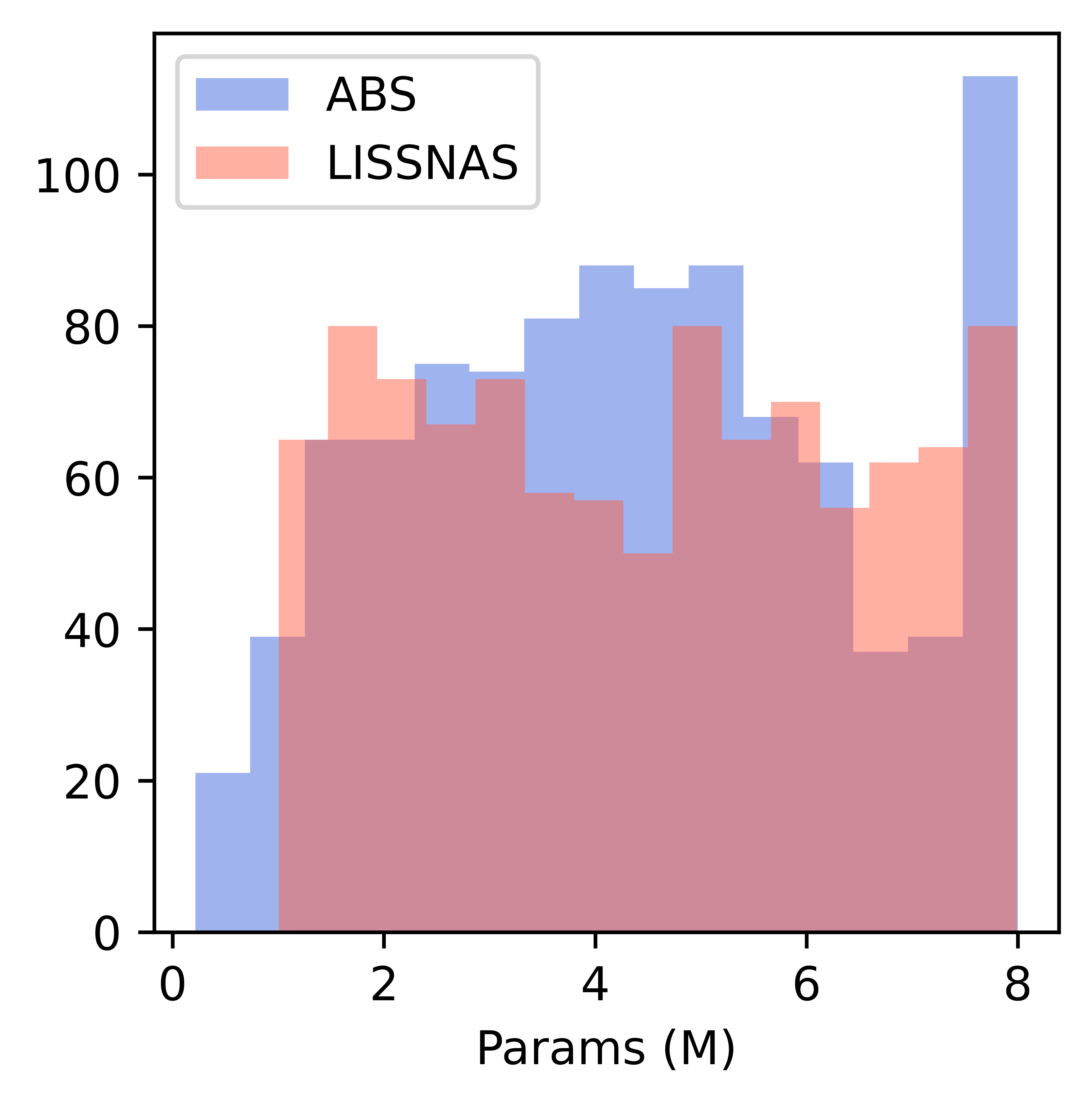}
      \label{fig:paramshist}
    \end{subfigure}
\end{center}
\caption{Histograms in the ShuffleNetV2 search space show bias for larger networks in other algorithms for larger models. More results are in the appendix.}
\label{fig:hist}
\end{figure}

From Figure \ref{fig:hist}, we can see that our search space does a better job of preserving networks with fewer FLOPs. Without special tuning, other shrinkage algorithms are incapable of preserving the smaller networks. Our method inherently preserves architectural diversity in contrast to being tuned to run in specific FLOP ranges. 

Since vector embeddings are common for architecture representations, we utilize embeddings to compute the maximum cosine distance between architectures in a search space. We utilize the same embedding scheme across all architectures for fair comparison. Our results in Table \ref{table:cosdis}, substantiate our previous diversity observations. From this standpoint, dissimilar networks have a higher cosine distance and indicate diversity within the space.  Interestingly, we also found that our shrunk search space is a negligible distance away from covering the complete original space. 

\begin{table}[htbp]
\begin{center}    
\begin{tabular}{lll}
\hline
\noalign{\smallskip}
\multirow{4}{*}{NASBench101}    & PADNAS         & 135.2             \\
                                & ABS               & 167.3             \\
                                & NSE               & 201.5             \\
                                & \textbf{LISSNAS}  & \textbf{255.6}    \\
                                & Original          & 260               \\
\hline \noalign{\smallskip}
\multirow{4}{*}{ShuffleNetV2}       & PADNAS             & 17.9              \\
                                    & ABS                   & 21.2              \\
                                    & NSE                   & 21.6              \\
                                    & \textbf{LISSNAS}      & \textbf{28.0}     \\
                                    & Original              & 30                \\

\hline
\end{tabular}
\caption{Maximum Cosine Distance }
\label{table:cosdis} 
\end{center}
\end{table}


%% file: conclusion.tex
In summary, we present a novel search space shrinkage method that improves NAS performance. Through exploiting locality, LISSNAS is able to maintain architectural diversity while improving quality as measured through a variety of statistical approaches. The {\it shrink index}, can be used to compare search space shrinkage works in the future. We believe that search space optimization holds much promise for driving future improvement in NAS.

%% file: ack.tex
This work was supported in part by the National Science Foundation through NSF grants CNS-2112562 and IIS-2140247. Special thanks to  Shakthi Visagan for the feedback and editing.

%% file: appendix.tex
\section{Appendix}

\section{Definitions, Instantiations and Clarifications} \label{dai}

\begin{itemize}
    \item Locality: Locality as defined in NASBench101 \cite{ying2019nasbench}, is the property that architectures that are similar in structure or composition tend to have similar performance metrics. 
    \item Search space: A search space consists of a set of neural architectures that are candidates for search. Synonymously, the search space defines which architectures can be represented. 
    \item Model Distribution: Network Design Spaces \cite{designing2019} proves and demonstrates the importance of using a distribution of models for the evaluation of a search space. Our definition of a model distribution is consistent with this work. We sample and evaluate a set of models from a search space giving rise to a model distribution. We evaluate the distribution using tools from classical statistics. 
    \item Edit Distance: The minimum number of changes needed to convert one architecture to another. Changes assume different forms for different search spaces. For more details, please see Appendix \ref{edtext}.
    \item Isomorphism: Isomorphisms are graphs that appear different but have the same underlying structure. Isomorphs have an edit distance of 0. In our work, we eliminate isomorphisms and only retain one network of a given structure within the space.
    \item Neighbors: We refer to networks that are a quantifiable number of changes away from each other as neighbors. Note that neighbors at relatively small edit distances are more structurally similar than neighbors at large edit distances. 
    \item Optimal, Well-performing: We will utilize the terms {\it well-performing} and {\it optimal} in addition to {\it poor-performing} and {\it suboptimal} interchangeably. 
\end{itemize}

\section{Search Space and respective dataset details} \label{dataset}

NASBench101 and NASBench301 are cell-based search spaces containing 423K and $10^{18}$ architectures respectively. They both represent networks as a series of stacked cells where the cell structure is represented by a Directed Acyclic Graph (DAG). ShuffleNetV2 is a block-based search spaces. It contains $4^{20}$ architectures. Block-based search spaces have a fixed macrostructure and contain design choices for each layer. TranNASBench is a combination of a block-based and cell-based search space containing 7K networks. In this search space, both the macro architecture and cell structure can be searched. These search spaces provide vastly different sizes and structures and span a wide range of vision tasks from object classification to semantic segmentation.

\subsection{NASBench101}
NASBench101 contains 423K unique architectures trained on CIFAR-10. Each model is trained multiple times resulting in several validation accuracies reported by the API. For the purpose of this work, we take the maximum reported validation accuracy for a network when querying the dataset. The macro architecture consists of 3 stacked blocks with 3 cells in each block. There is a convolutional stem and downsampling between each block. The cell structure consists of up to 5 nodes and 9 edges with operations lying in the nodes. Each node can take one of 3 choices: 3x3 convolution, 1x1 convolution, and 3x3 max pool. More details on the dataset can be found in the original paper here \cite{ying2019nasbench}.

\subsection{ShuffleNetV2}
The ShuffleNetV2 search space details are outlined by SPOS \cite{guo2019singlepath}. In summary, the search space contains $4^{20}$ architectures. We elect to model this dataset using a supernet trained on ImageNet. The architectures in the space share the same macrostructure, a linear stack of choice blocks with a convolutional stem and a fully connected output layer. There are 20 choice blocks that can take one of 4 different operations: choice\_3, choice\_5, choice\_7, and choice\_x that differ in kernel size and the number of depthwise convolutions.

\subsection{TransNASBench}
TransNASBench consists of 7,352 architectures with changes at both the macro-level and cell-level. This dataset contains object classification, scene classification, room layout, jigsaw, autoencoding, surface normal, and semantic segmentation \cite{transNASBench}.

\section{Edit Distance} \label{edtext}
\begin{figure}[h]
    \centering
    \includegraphics[width=\linewidth]{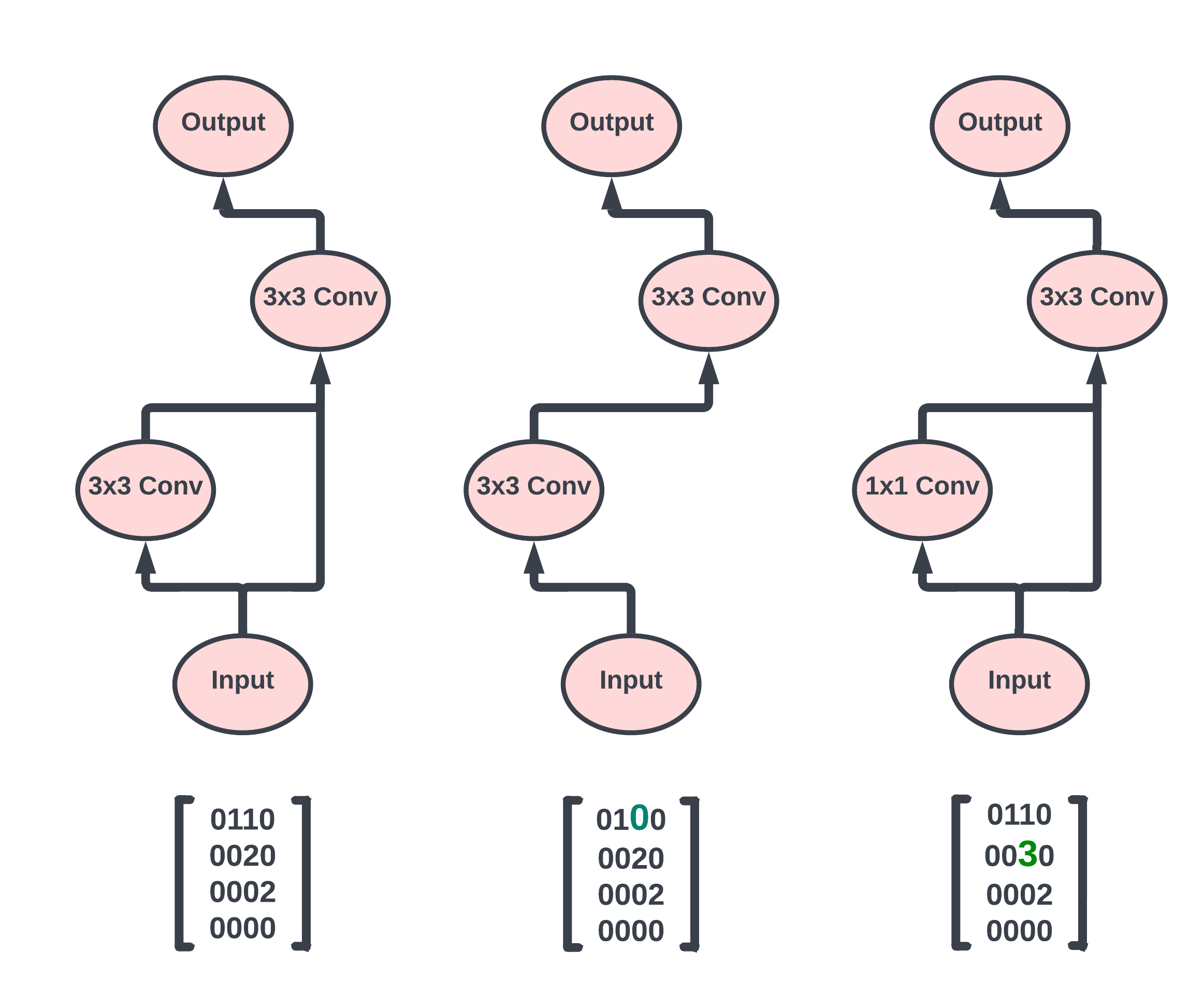}
    \caption{Edit distance example: The original adjacency matrix is presented on the left. Note that input "operation" is denoted by a 1 in the matrix, 3x3 convs denoted by 2, and 1x1 convs denoted by 3. The middle matrix shows the removal of an edge. The right matrix shows the change of an operation. Both of these changes are at edit distance 1 from the original network.}
    \label{fig:ed}
\end{figure}

We can represent architectures in either adjacency matrices or choice lists for cell-based and block-based search space respectively. Adjacency matrices are square matrices with the number of rows and columns representing the number of nodes in the graph. Edges are represented as 1s and lack of edges is represented by 0s. In order to embed operation choices in these graphs we choose to replace the 1 with an encoding for the operation choice. As an example in NASBench101, a 1 represents a 3x3 conv, a 2 represents a 1x1 conv, and a 3 represents a 3x3 max pool. Figure \ref{fig:ed} shows the adjacency matrix and corresponding DAG structure in addition to 2 examples of edit distance of 1. We follow a similar definition for edit distance in the block-based search spaces. For example, in ShuffleNetV2, operation choices are encoded using the numbers 0-3 with each representing a particular operation. A 20-element vector can then represent an architecture in this space. These types of embedding schemes are commonly used in NAS work.

\section{Edit Distance Additional Results}
\begin{figure}[h]
    \centering
    \includegraphics[width=\linewidth]{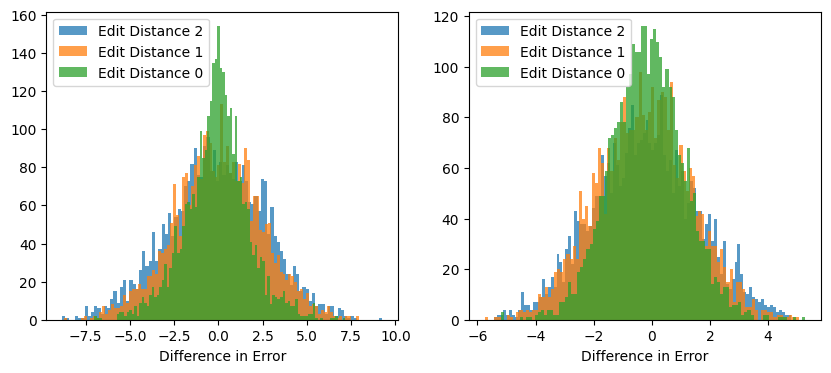}
    \caption{Histograms show the difference in accuracy between networks as edit distance increases. Scene classification (left) and object classification (right) follow the same trend as all tasks in TransNASBench. }
    \label{fig:historgrams}
\end{figure}

\begin{table}[h]
\begin{center}  
\begin{tabular}{lll}
\hline
\noalign{\smallskip}
Edit Distance   & Type of Change        & AAD(\%)               \\
\multirow{4}{*}{1}  & Both              & 0.83               \\
                    & Operation         & 0.85               \\
                    & Edge Add/Remove   & 0.80      \\
\hline \noalign{\smallskip}
\multirow{4}{*}{2}  & Both              & 1.75               \\
                    & Operation         & 1.80               \\
                    & Edge Add/Remove   & 1.74      \\
\hline \noalign{\smallskip}
\multirow{4}{*}{3}  & Both              & 2.40               \\
                    & Operation         & 2.42               \\
                    & Edge Add/Remove   & 2.39      \\
\hline
\end{tabular}
\caption{Type of Change Impact in NASBench101}
\label{table:ed2}   
\end{center}
\end{table}

\section{Supernetwork Training Details}
On the NASBench101, we use the same training scheme as Ma et al., \cite{ma2018shufflenet} (including data augmentation, learning rate scheduling, etc.) for the supernet and the final architecture. For the supernet, we use a batch size of 1024 and a ghost batch size of 128 and then train for 120 epochs in a distributed fashion on 8 Titan XP GPUs. We train the final architecture on a single A5000 GPU with a batch size of 256 and a learning rate of 0.0125.

For the ShuffleNetV2, we train our supernet for 200 epochs, using a batch size of 64 and a ghost batch size of 32. We incorporate standard initial learning rate, learning rate scheduling, and momentum in addition to the standard data augmentation techniques used on the CIFAR-10 dataset. \cite{krizhevsky2009cifar10}. For the baseline one-shot network on the full search space, we follow a balanced path sampling strategy used by \cite{luo2020balanced} on a single Titan XP GPU. 

Lastly, we train an XGBoost predictor \cite{Chen_2016} on a validation set of 10K network-accuracy pairs to run our algorithm.

Finally, we retrained the best 10 networks from LISSNAS search to obtain the single best architecture.

\section{Understanding Assumptions} \label{ass}
\begin{itemize}
    \item In all of our search spaces, we assume that every change that can be made to a network is of equal value. Effectively, at the same edit distance, the accuracy difference between a generated neighbor and the original network is roughly the same, irrespective of the type of change. We understand that this assumption is likely incorrect as changing an operation has a greater impact than changing an edge. However, this served as a reasonable and effective heuristic for our algorithm. As shown in Table \ref{table:ed2}, the AAD from both types of changes is similar. Examples of potential changes are demonstrated in Figure \ref{fig:ed}.
    \item For all of our data analysis we consider a minimum of 1000 samples. While DDS provides enough justification for this assumption \cite{designing2019,radosavovic2020designing}, we still experimentally verified across all of our search spaces and datasets. We also note that we conducted our verification experiment several times for each of our datasets and search spaces. Our experiments entailed randomly sampling 1000 nets from the original space. We then determined that the distributions of the sampled space and the entire search space were the same using a variety of measures including the error EDF. Additionally, our future work involves determining the minimum threshold of networks from each search space that need to be initially sampled to represent the entire distribution of the search space. Computing this threshold could greatly contribute improve efficiency.
\end{itemize}


\section{Ablation Studies}\label{ablation}
For a more in-depth understanding of our technique, we address the following questions through ablation studies:
\begin{enumerate}
\item What is the performance improvement between our approach and a naive random shrink? A naive random shrink involves randomly sampling an existing search space for a predefined number of samples. 
\item Why do we exploit locality to generate neighbors? i.e. if we retain the best seeds from each iteration and do not generate neighbors from these seeds, do we still get good results? 
\item Instead of using locality to generate neighbors, can we query all networks in the search space using the predictor? 
\end{enumerate}
\subsection{Naive Random Shrinkage}
\begin{table}[h]
\begin{center}
\begin{tabular}{lcc}
\hline\noalign{\smallskip}
Search Space & Avg Acc & Max Acc  \\
\noalign{\smallskip}
\hline
\noalign{\smallskip}
Full Search Space  &  90.4  &  94.6  \\
Top-30  &  90.8  &  93.2 \\
Top-20  &  91.0  &  93.5 \\
Top-10  &  91.4  &  94.1 \\
Top-5  &  91.9  &  93.9 \\
{\bf LISSNAS} &  92.1  &  94.6  \\
\hline
\end{tabular}
\caption{Naive Random}
\label{table:naive}
\end{center}
\end{table}

In this section, we compare our algorithm to a naive algorithm that randomly samples the space and preserves the top-x\% of networks. Table \ref{table:naive} shows that the maximum accuracy of random approaches suffers from randomness. The Top-5 random shrinkage approach has a lower max Top1 accuracy that the Top-10 shrinkage. All of the shrinkage methods were given the same query budget for a fair comparison. 

\subsection{Neighbor Generation Methods}

\begin{table}[h]
\begin{center}
\begin{tabular}{lcc}
\hline\noalign{\smallskip}
Technique & \# of Queries & Top1  \\
\noalign{\smallskip}
\hline
\noalign{\smallskip}
No Neighbor &  50K  &  93.9 \\
Without Locality &  196K  &  93.2 \\
With Locality  &  100K  &  94.6 \\
\hline
\end{tabular}
\caption{NASBench101 Neighbor Generation}
\label{table:ab1}
\end{center}
\end{table}

The first line in Table \ref{table:ab1} represents not using locality or any form of neighbor generation. This mirrors the best single iteration results from our algorithm and the best results from naive random shrinkage. For the second line, we used the same overall structure as LISSNAS but instead of generating neighbors using locality, we randomly sample networks to add back into the next iteration. This technique produced much more variance with premature sub-optimal termination or excessive iterations with slow convergence. The final line represents the performance achieved by our algorithm. We can see from Table \ref{table:ab1} that locality performs significantly better than random sampling for neighbor generation and converged faster. 
\subsection{Query All Networks}
While querying all networks might be feasible in small search spaces, employing this method on modern large search spaces quickly proves to be computationally infeasible. Assume we are able to obtain a predictor for the FairNAS search space containing $6^{19}$ architectures. Assuming each query to the predictor takes 1ms, querying every network in the space would take in the neighborhood of 27 million years. Through LISSNAS, we exploit locality to significantly reduce the number of networks we need to query in order to shrink the space and search in a reasonable amount of time.


\section{EDF Additional Results}
This figure is the continuation of the results presented in Section \ref{sss}.
\begin{figure}[h]
    \centering
    \includegraphics[width=\linewidth]{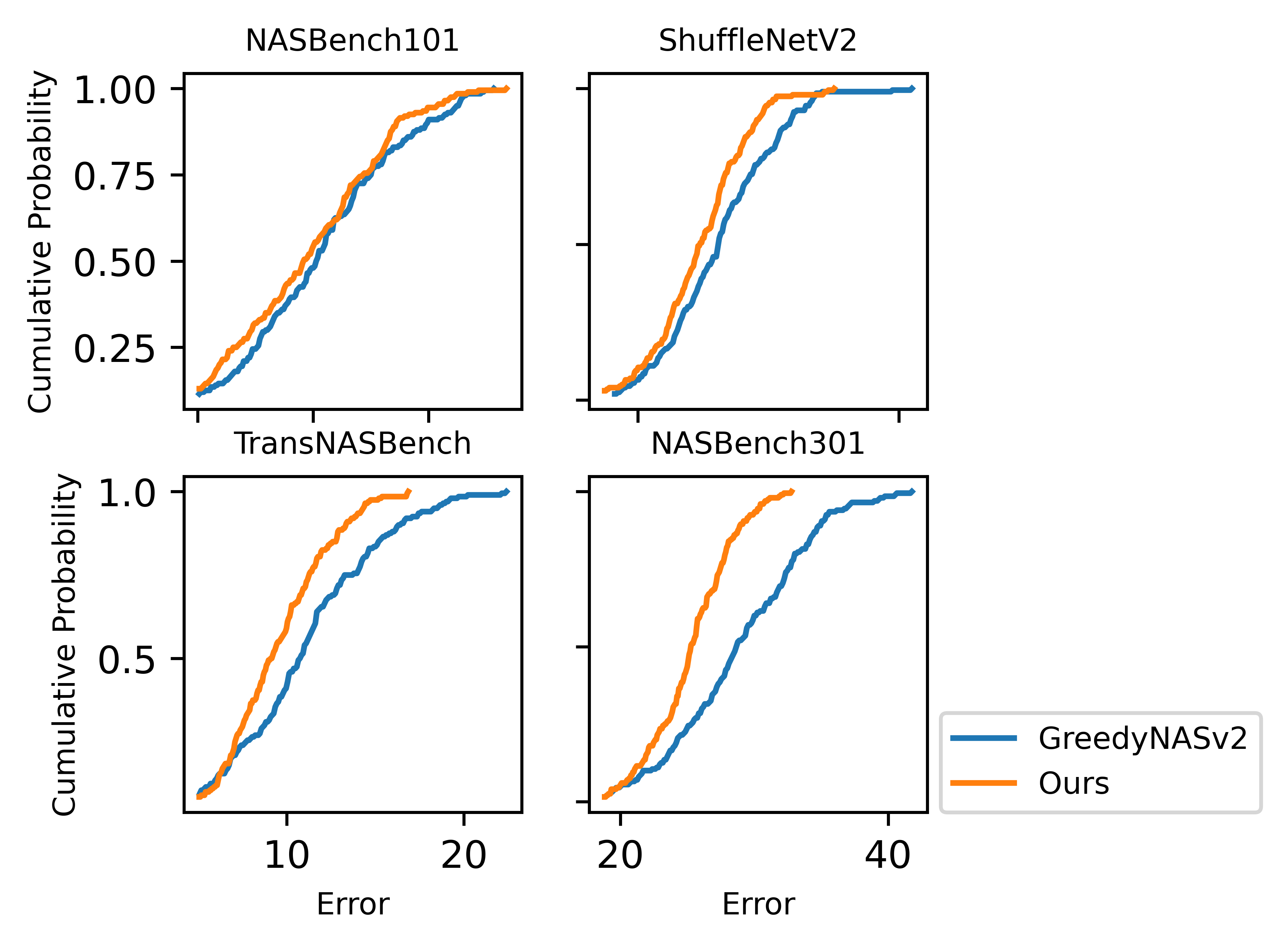}
    \caption{Error EDF captures the quality of a search space. Our algorithm performs significantly better than GreedyNASv2 in the larger search spaces.}
    \label{fig:edfGreedy}
\end{figure}

\section{Cosine Distance Additional Results}
This table is the continuation of the results presented in Section \ref{ad}.
\begin{table}[h]
\begin{center}   
\begin{tabular}{lll}
\hline
\noalign{\smallskip}
\multirow{4}{*}{TransNASBench}   & PADNAS         & 5.3               \\
                                        & ABS               & 4.2               \\
                                        & NSE               & 5.7               \\
                                        & \textbf{LISSNAS}  & \textbf{8.9}      \\
                                        & Original          & 10                \\

\hline \noalign{\smallskip}
\multirow{4}{*}{NASBench301}    & PADNAS         & 673.2             \\
                                & ABS               & 612.7             \\
                                & NSE               & 619.9             \\
                                & \textbf{LISSNAS}  & \textbf{845.1}    \\
                                & Original          & 947               \\
\hline
\end{tabular}
\caption{Maximum Cosine Distance }
\label{table:cosdis2}  
\end{center}
\end{table}